\documentclass[sigconf]{aamas}

\usepackage{graphicx}
\usepackage{balance}
\usepackage{comment}
\usepackage{fixltx2e}
\usepackage{amsmath}
\usepackage{balance}
\usepackage{comment}
\usepackage{subfigure}
\usepackage{epsfig}
\usepackage{url}
\usepackage{amsfonts}
\usepackage{mdwmath}
\usepackage{mdwtab}
\usepackage{cite}
\usepackage{booktabs}
\usepackage{tabularx}
\usepackage{multirow} 
\usepackage[]{caption}
\usepackage{amsmath}
\usepackage{hyperref}
\usepackage[fleqn]{mathtools}
\usepackage{natbib}
\usepackage{graphicx}
\usepackage{relsize}
\usepackage{flushend}
\usepackage{booktabs}
\usepackage{adjustbox}
\usepackage[linesnumbered,ruled,vlined]{algorithm2e}
\usepackage{amsmath}
\usepackage{breqn}

\setcopyright{none}
\acmConference[ALA '22]{Proc.\@ of the Adaptive and Learning Agents Workshop (ALA 2022)}
{May 9-10, 2022}{Online, \url{https://ala2022.github.io/}}{Cruz, Hayes, da Silva, Santos (eds.)}
\copyrightyear{2022}
\acmYear{2022}
\acmDOI{}
\acmPrice{}
\acmISBN{}
\acmSubmissionID{33}

\title[]{A Reinforcement Learning Approach for Electric Vehicle Routing Problem with Vehicle-to-Grid Supply}

\author{\Large{Ajay Narayanan, Prasant Misra, Ankush Ojha, Vivek Bandhu$^{\#}$, Supratim Ghosh, Arunchandar Vasan}}
\affiliation{
  \institution{Tata Consultancy Services - Research | $^{\#}$Tata Consultancy Services}
  \city{}
  \country{}
}
\email{[ajay.narayanan, prasant.misra, ojha.ankush, vivek.bandhu, supratim.ghosh2, arun.vasan]@tcs.com}

\begin{abstract}
The use of electric vehicles (EV) in the last mile is appealing from both sustainability and operational cost perspectives. In addition to the inherent cost efficiency of EVs, selling energy back to the grid during peak grid demand, is a potential source of additional revenue to a fleet operator. To achieve this, EVs have to be at specific locations (discharge points) during specific points in time (peak period), even while meeting their core purpose of delivering goods to customers. 
In this work, we consider the problem of EV routing with constraints on loading capacity; time window; vehicle-to-grid energy supply (CEVRPTW-D); which not only satisfy multiple system objectives, but also scale efficiently to large problem sizes involving hundreds of customers and discharge stations.
We present \quik\ that uses reinforcement learning (RL) for EV routing to overcome these challenges.
Using Solomon datasets, results from RL are compared against exact formulations based on mixed-integer linear program (MILP) and genetic algorithm (GA) metaheuristics.
On an average, the results show that RL is $24$ times faster than MILP and GA, while being close in quality (within $20$\%) to the optimal.
\end{abstract}


\newcommand{\quik}{\textit{QuikRouteFinder}}

\begin{document}

\pagestyle{fancy}
\fancyhead{}

\maketitle

\section{Introduction} \label{sec:introduction}

Last-mile delivery is the most expensive part of the logistic and e-commerce process, often accounting for more than $50$\% of the overall shipping expenses~\citep{businessinsider2022:lastmile}.
To address this major pain-point, most players in this space are exploring the use of newer technologies that can help in reducing the cost. 
Electric vehicles (EV) are a reasonable choice for this application, as they offer many cost benefits that come naturally from electrification such as high fuel efficiency; low maintenance overhead; compliance to emission regulations.
Therefore, there is an increasing adoption of EVs in last-mile delivery operations; expected to grow to $8$ million by $2030$~\citep{mckinsey2020:evfleets,misra2021:evmobility}. 
While these developments are promising, the benefit of using last-mile delivery EVs extend beyond this present scope.
In fact, they can be used to provide other ancillary services within the city ecosystem.
Such multi-service operation models open up alternate revenue channels for the logistics provider, which can lead to further cost reduction of the last-mile segment. 
Motivated by this observation, we explore an alternate delivery model using EVs; where the scope is not only limited to delivery of goods to customers, but extends to the delivery (or selling) of energy to the grid during its periods of peak energy demand. Due to the virtue of the on-board battery (which provide fuel to move the vehicle) EVs can potentially use the same energy reserve to sell power back to the grid at higher tariff rates.
The key question, however, is whether or not last-mile EVs can perform this additional task without deviating from the primary business objective.
\newline
\indent
We approach the problem of delivery of goods (to customers) and delivery of energy (to the grid through discharging stations) as an electric vehicle routing problem (VRP) with the goal of finding the most cost effective routes that guarantee order fulfillment.
This problem is non-trivial for the following two reasons.
First, the delivery demand (i.e., goods/energy) is present at specific time periods and locations, and therefore, the supply (i.e., delivery EVs) needs to match that demand (both in space and time) for making a successful transaction.
Second, the computationally difficult VRP problem becomes even harder with the introduction of new constraints related to EVs and energy discharge (that will happen both implicitly due to travel and explicitly due to energy sale at discharging stations).
A number of solutions exist for solving VRP and its different variations.
They range from mixed-integer linear programs (MILP) providing exact solutions to metaheuristics involving either neighborhood search, tabu search, genetic algorithms, simulated annealing, or ant colony optimization that provide near exact solutions with faster convergence~\citep{erdelic2019:evrp-survey}. 
However, none of these approaches scale to large problem instances that involve hundreds of customers and discharging stations with multiple operational constraints in quick time~\citep{sultana2021:rl-vrp,gupta2022:rl-vrp} with reasonable solution accuracy.
\indent
We present \quik\ that uses reinforcement learning (RL) to address the problem of routing for EVs with multi-service delivery by overcoming the above challenges.
In RL, a controller agent observes a system state and decides a control action. 
The system implements the action and provides a reward to the agent. 
In our case, the RL agent acts as an aggregator to a fleet of delivery EVs. 
The state of the system at every time step includes information about individual (vehicle-customer) or (vehicle-discharging station) pairs.
The action taken is to decide the visit location of vehicles in the system.
The reward to the agent would be the trip cost obtained in that timestep of the control action.
The RL agent algorithmically \emph{learns} to map actions to the observed system state in a way that maximizes the rewards earned over the training period.
This offline training helps the agent to make faster decisions in real-time with efficient scaling to large problem sizes.
\vspace{1mm}
\newline
\noindent
In this paper, our specific contributions include the following.
\vspace{1mm}
\newline
\noindent
$(1)$ We model the electric vehicle routing problem (CEVRPTW-D) with constraints on loading capacity; time window; vehicle-to-grid energy supply; and formulate the multi-objective optimization problem to minimize the trip cost of the fleet.
\vspace{1mm}
\newline
\noindent
$(2)$ We design \quik\ : a value-based reinforcement learning algorithm by defining the (state, action) space, and engineer the reward signal for the RL agent to find the cost effective delivery routes.
We design and implement a genetic algorithm (GA) metaheuristic to derive optimal results for CEVRPTW-D.
\vspace{1mm}
\newline
\noindent
$(3)$ Using Solomon datasets, we evaluate and compare the computation speed and solution accuracy of the proposed RL model against GA and MILP. 
Preliminary results from our study show that \quik\ (based on RL) is $24$ times faster than the GA and MILP baselines in terms of solutioning speed, but with $\approx 20$\% decrease in solution quality.
\newline
\indent
The rest of the paper is organized as follows.
The system overview is outlined in Section~\ref{sec:system}.
Section~\ref{sec:evaluation} presents the evaluation studies.
We review existing literature in Section~\ref{sec:relatedwork}, and conclude by summarizing the work in Section~\ref{sec:conclusion}.

\section{System Overview} \label{sec:system}

The last-mile logistics system for CEVRPTW-D consists of four primary stakeholders.
First, the goods demand-side, consisting of customers who are expecting the delivery of ordered items within a certain time window.
Second, the energy demand-side, consisting of discharging stations willing to purchase power during peak energy demand periods of the grid.
Third, the supply-side, comprising of last-mile delivery EVs.
The final stakeholder is the fleet manager that connects the demand and the supply; but in a manner that ensures that the demand for goods is met without failure, and demand for energy is served if there is an opportunity.
Our perspective is that of the fleet manager with an aim to minimize trip cost of the fleet by finding cost effective routes; which reduces the overall travel distance and vehicles used for delivery, while generating revenue from energy sale to the grid. 
\newline
\indent
The fleet manager needs to plan the delivery schedule and vehicle routes in advance.
All requisite information is known a priori since the fulfilment center shares the customer order list for delivery purpose, while the grid announces its peak energy demand periods, normally, a day before the requirement.
We consider an operational setup where a fleet of last-mile EVs leave the base depot at a given time with a starting state-of-charge (SoC) of $Q$.
Their target is to complete all deliveries and return back to the same depot.
During this delivery period, EVs could also visit a discharging station and sell power back to the grid in a time defined manner that does not impact the delivery of customer orders.
We assume that $Q$ is sufficient to complete the delivery trip without recharging and, in addition, may allow V$2$G discharging where possible. 
This setup does not allow recharging halts during the trip as it would increase the vehicle idle time, and subsequently, the trip cost.
\vspace{1mm}
\newline
\noindent
\textbf{System model:} Let $G = (V,E)$ represent a complete undirected graph; where the set of nodes is denoted by $V = \{v_{0}\} \cup K \cup P$, and the set of edges connecting the nodes is denoted by $E$ = $\{(i,j): i,j \in V, i \neq j\}$.
The set of nodes has three subsets: $v_{0}$ denotes the depot; $K = \{k_{1},...,k_{m}\}$ denotes the set of customers; $P = \{p_{1},...,p_{n}\}$ denotes the set of discharging stations.
The depot $v_{0}$ holds a homogeneous fleet of EVs denoted by $X = \{x_{1},...,x_{u}\}$, each with a carrying capacity of $C$ and starting SoC of $Q$. 
Each customer $k_{m} \in K$ has a positive demand of $c_{i}$; service time of $s_{i}$; and time window of [$e_{i}$, $l_{i}$].
Each edge is associated with a travel distance of $d_{ij}$; travel time of $t_{ij}$; and travel energy of $b_{ij}$.
The battery of the EV is depleted at $H$ (kWh/km) on the road (while traveling between two nodes); and $R$ (kWh) during the discharge operation (while giving back power back to the grid during the peak demand time periods between $G^{1}_{i}$ and $G^{2}_{i}$ at discharging station node $i$) (see Table~\ref{tab:vardef}).
\begin{table}[t]
\vspace{-2mm}
\caption{Variable definitions}
\vspace{-2mm}
\label{tab:vardef}
\begin{small}
\begin{tabular}{l l l}
\toprule
\textbf{Symbol}  & \textbf{Meaning}\\ 
\midrule
$t$  & time instant when decisions are made \\
$[1,T]$ & decision horizon \\
\midrule
\multicolumn{2}{l}{\bf Sets}          \\
\midrule
$V$ = \{$v_{o}\} \cup K \cup P$  & set of \{$m+n+1$\} nodes \\
$K = \{k_{1},...,k_{m}\}$ & set of $m$ customers \\
$P = \{p_{1},...,p_{n}\}$ & set of $n$ VPP stations \\
$X = \{x_{1},...,x_{u}\}$ & set of $u$ EVs \\
\midrule
\multicolumn{2}{l}{\bf Customer}          \\
\midrule
$e_{i}$ & earliest start of service at node $i$ (time)\\
$l_{i}$ & latest start of service at node $i$ (time)\\
$s_{i}$ & service time at customer node $i$ \\
$c_{i}$ & demand (of goods) at customer node $i$ \\
\midrule
\multicolumn{2}{l}{\bf Travel}          \\
\midrule
$d_{ij}$ & distance between nodes $i$ and $j$ \\
$t_{ij}$ & travel time between nodes $i$ and $j$ \\
$H$ & charge consumption (kWh/km) rate\\
$b_{ij}$ & energy consumed in travelling between \\
         & nodes $i$ and $j$ (=$H.d_{ij}$) \\
\midrule
\multicolumn{2}{l}{\bf Vehicle}          \\
\midrule
$C$ & carrying capacity of each EV \\
$Q$ & starting state-of-charge (SoC) of each EV \\
\midrule
\multicolumn{2}{l}{\bf Discharging Station \& Grid}          \\
\midrule
$R$ & discharging rate of each discharging station \\
$G^{1}_{i}$ & start time - grid peak demand at station node $i$ \\
$G^{2}_{i}$ & stop time - grid peak demand at station node $i$ \\
\midrule
\multicolumn{2}{l}{\bf Decision Variables} \\
\midrule
$\alpha_{ij}$ & indicates, if edge $ij$ is traversed by an EV (binary) \\
$\gamma_{i}$ & service time at discharging station node $i$ \\
$\tau_{i}$ & time-of-arrival at node $i$ \\
$\theta_{i}$ & remaining battery capacity on arrival at node $i$\\
$\lambda_{i}$ & remaining cargo on arrival at node $i$ \\
\bottomrule
\end{tabular}
\end{small}
\vspace{-2mm}
\end{table}

\begin{figure*}[t]
\begin{center}
\includegraphics[width=7in]{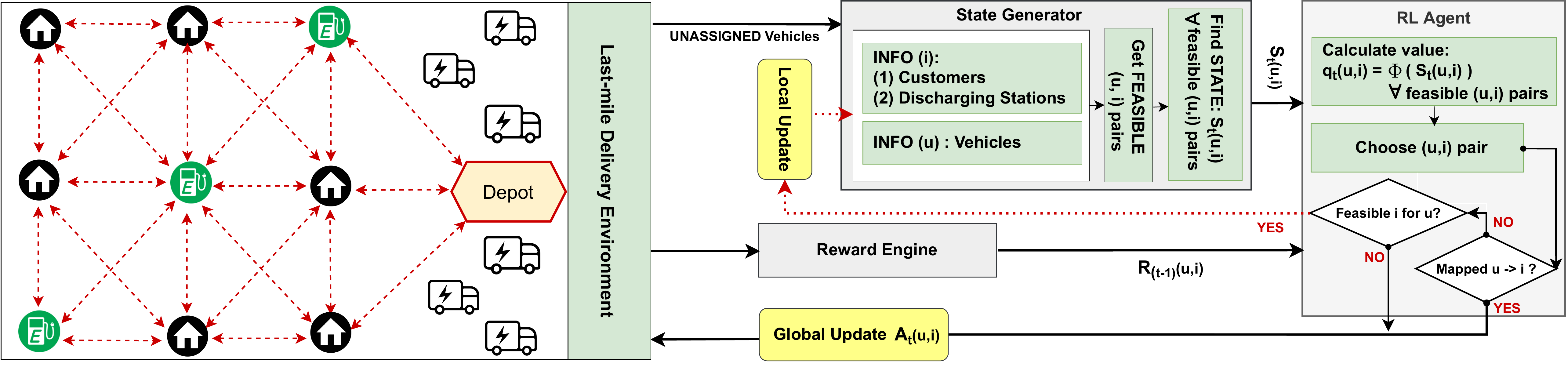}
\end{center}
\caption{System model of QuikRouteFinder.}
\label{fig:system-model}
\end{figure*}

\subsection{Optimization Problem} \label{sec:milp}

The objective of CEVRPTW-D is to minimize the trip cost:
\begin{equation}\label{milp:obj}
    \mathcal{M} = \min \left(Y_{1}.\sum d_{ij}\alpha_{ij}  + Y_{2}.\sum \alpha_{0i} - Y_{3}.\sum \gamma_{i} \right)
\end{equation}
$\mathcal{M}$ is taken as the combined weighted\footnote{The weights $Y_{1} = 0.0354$, $Y_{2} = 101.81$, $Y_{3} = 0.2478$ are adjusted according to EU regulations~\citep{leaseplan:ev}.} cost of the total distance travelled by the EVs and the number of such EVs used in a trip; minus the reward gained due to the time spent at discharging stations and selling power back to the grid.
\vspace{1mm}
\newline
\noindent
Eq.~(\ref{milp:obj}) is subject to the following constraints:
\begin{equation}
 \sum_{j \in V,i \neq j}\alpha_{ij} = 1 \hspace{48mm} \forall i \in K, \label{eqn:array1} \\
\end{equation}          
\begin{equation}
 \sum_{j \in V,i \neq j}\alpha_{ij} = \sum_{j \in V,i \neq j}\alpha_{ji} \hspace{36mm} \forall i \in V,  \label{eqn:array2} \\
\end{equation}
\begin{equation}
\tau_{i} + (t_{ij} + s_{i})\alpha_{ij} - l_{0}(1-\alpha_{ij}) \leq \tau_{j} \hspace{10mm} \forall i \in K,\forall j \in V,i \neq j,\label{eqn:array3}\\
\end{equation}
\begin{equation}
\tau_{i} + (t_{ij} + \gamma_{i})\alpha_{ij} - l_{0}(1-\alpha_{ij}) \leq \tau_{j} \hspace{10mm} \forall i \in P,\forall j \in V,i \neq j,\label{eqn:array4}\\
\end{equation}
\begin{equation}
e_{i} \leq \tau_{i} \leq l_{i} \hspace{53mm} \forall i \in K, \label{eqn:array5}\\
\end{equation}
\begin{equation}
0 \leq \theta_{j} \leq \theta_{i} - (H.d_{ij})\alpha_{ij} + Q(1-\alpha_{ij}) \hspace{5mm} \forall i \in K,\forall j \in V,i \neq j,\label{eqn:array6}\\
\end{equation}
\begin{equation}
0 \leq \theta_{j} \leq \theta_{i} - (H.d_{ij} + R.\gamma_{i})\alpha_{ij} + Q(1-\alpha_{ij}) \forall i \in P,\forall j \in V,i \neq j,\label{eqn:array7}\\
\end{equation}
\begin{equation}
0 \leq \lambda_{j} \leq \lambda_{i} - c_{i}\alpha_{ij} + C(1-\alpha_{ij}) \hspace{25mm} \forall i,j \in V,\label{eqn:array8}\\
\end{equation}
\begin{equation}
\tau_{i} \geq G^{1}_{1} \hspace{49mm} \forall i \in P, \tau_{i} \neq 0,  \label{eqn:array9} \\
\end{equation}
\begin{equation}
\tau_{i} + \gamma_{i} \leq G^{2}_{i} \hspace{44mm} \forall i \in P,\tau_{i}\neq 0.   \label{eqn:array10} \\
\end{equation}

\vspace{1mm}
\noindent
Constraint~(\ref{eqn:array1}) ensures that every customer is visited exactly once, while making it optional to visit any of the discharging stations.
Constraint~(\ref{eqn:array2}) establishes flow conservation, wherein the the number of incoming arcs is equal to the number of outgoing arcs at each node.
Constraints~(\ref{eqn:array3}) and (\ref{eqn:array4}) guarantee time feasibility of arcs leaving customers, discharging stations, and the depot.
The condition that each node must be visited within its time window is ensured by 
constraint~(\ref{eqn:array5}).
Constraints~(\ref{eqn:array6}) and (\ref{eqn:array7}) enforce remaining charge (energy) feasibility for arcs leaving customers, discharging stations, and the depot.
The demand fulfillment of all customers awaiting order delivery is guaranteed by constraint~(\ref{eqn:array8}).
Constraints~(\ref{eqn:array9}) and (\ref{eqn:array10}) ensure that the discharge service time aligns with the grid peak demand period, if discharging stations are visited.
Note that all nodes are given a time window ranging between the starting time of service and the total time horizon.
Therefore, by ensuring that the time window constraints for individual customers and discharging stations are met, it can also be ensured that the entire operation happens within the specified time horizon.

\subsection{RL Representation}

The optimization problem is formulated in the RL framework by defining the (state, action) space and engineering the reward function. 
The algorithm builds the solution by mapping vehicles to a (customer or discharging station) node at every decision epoch, eventually leading to an optimized route. 
\vspace{2mm}
\newline
\noindent
\textbf{States:} 
The information about individual (vehicle-node) pairs are given as input to the RL agent.
The definition used for the observed system state $S_{t}(u,i)$ (for vehicle $u$; at time $t$; for proposed node $i$) is shown in Table~\ref{tab:rlstate}.
In our system model, vehicles spend energy under two conditions: first, if they are  traveling between two nodes; and second, if they are participating in energy (discharge) transactions at discharging stations.
These two aspects are captured by $b_{i,j}$ and $z_{i}$, respectively.
$b_{i,j}$ also serves as a proxy variable to ascertain the distance travelled between nodes.
There are two supporting binary variables $I_{depo}$ and $I_{cust}$ that, respectively, help to capture if the vehicle has left the depot, and the location of the proposed node (customer or discharging station).
The final variable $w_{u}^{i}$ captures the wait time of a vehicle $u$ at the proposed node $i$ before it can start service.
The following constants are used for normalizing non-binary state variables: (i) $E$ (that denotes the energy required to travel the diagonal length of the graph) is used for $b_{i,j}$ and $z_{i}$; (ii) $T$ (that denotes the decision time horizon) is used for $w_{u}^{i}$.
\vspace{2mm}
\newline
\noindent
\textbf{Action:} The action $A_{t}(u,i)$ is the (vehicle, node) pair at decision time $t$. 
It is selected either at the start of an episode when all vehicles are available, or when any vehicle becomes available post completing the assigned service.
This action is derived in the following manner.
The agent computes the respective value for each $(u,i)$ pair (irrespective of their availability status), and chooses the pair with the maximum value.
If the chosen ($u \mapsto i$) mapping includes a vehicle that is currently in-service, a local state update is made to capture this assignment.
This update includes removing the chosen node from the service list, and making changes to the distance and time factors.
The process is repeated either until all available vehicles or nodes get assigned;  post which the global environment is updated along with the reward for the assignment action.
\vspace{1mm}
\newline
\noindent
\textbf{Reward:} For each chosen (vehicle, node) pair, the reward $R_{t}(u,i)$ is taken as the total cost of vehicle $u$ visiting node $i$ at time $t$. It is defined as:
\vspace{-1mm}
\begin{equation}
R_{t} = -A_{1}*b_{ij} + A_{2}*z_{i} + A_{3}*I_{cust} - A_{4}*w_{u}^{i} - A_{5}*I_{depo}
\vspace{-1mm}
\end{equation}
Here, a positive reward is given for visiting either a customer node or a discharging station node; while a negative reward is given for choosing longer route segments; for assignments that lead to waiting time at nodes; and for sending new vehicles from the depot. 
We chose $A_{1} = 0.15$, $A_{2} = 0.001$, $A_{3} = 0.15$, $A_{4} = 0.15$, $A_{5} = 0.55$ since they gave the best solutions on the training data. 
In the reward function, since the weight $A_{3}$ (given for serving a customer node) is much greater than the weight $A_{2}$ (given for discharging station visit), the agent learns to prefer customer service more than discharge operation whenever presented with an option of both. 
In a bid to maximize the reward received, this preference of the agent of putting customers over discharge operation can be seen in the learning curve as the fulfilment ratio slowly becoming equal to $1$ (indicating that all the customers were serviced).
\vspace{2mm}
\newline
\noindent
\textbf{Handling workflow:} 
Once the reward is given for the corresponding (state, action) pair, the tuple [$S_{t}(u,i)$; $R_{t}(u,i)$; $q_{t}(u,i)$] is added to the replay buffer.
At the end of each episode, $\beta$ samples are drawn randomly from the replay buffer $B$; and the value network weights $\phi$ are updated by minimizing the mean squared error (MSE) between $q_{t}(u,i)$ and $R_{t}(u,i)$.
This makes the agent learn the (state, action) mapping.
\begin{table}[t]
\caption{RL State variables}
\vspace{-2mm}
\label{tab:rlstate}
\begin{small}
\begin{tabular}{l l}
\toprule
\textbf{Input} & \textbf{Explanation}\\ 
\midrule
$b_{i,j}$ & energy consumed in travelling between nodes $i$ and $j$ \\
$z_{i}$ & energy spent at node $i$ ($0$, if at customer; else $z_{i}$) \\
$I_{depo}$ & flag: indicates if vehicle $is$ starting from the depot \\
$I_{cust}$ & flag: indicates if node $i$ is a customer \\
$w_{u}^{i}$ & wait time of vehicle $u$ at node $i$ before it can start service \\
\bottomrule
\end{tabular}
\end{small}
\vspace{-5mm}
\end{table}
\vspace{2mm}
\newline
\noindent
\textbf{(Vehicle $\mapsto$ Node) masking scheme:} At each decision step $t$, the RL agent produces a scalar output for all feasible $(u,i)$ pairs.
We designed a masking scheme to derive these pairs.
Let us suppose that proposed next node of vehicle $u$, currently at node $i$, is node $j$. 
In such a case, pair ($u$, $j$) is considered \emph{infeasible} if it satisfies any of the following conditions:
\vspace{1mm}
\noindent
\newline
$\bullet$ Node $j$ is a customer with unfulfilled demand that is either nil or exceeds the remaining carrying load of vehicle $u$ (refer Eq.~(\ref{eqn:array8})).
\vspace{1mm}
\noindent
\newline
$\bullet$ Node $j$ is a customer and the current SoC of the vehicle cannot support the complete trip from node $i$ to node $j$ and back to the depot (refer Eq.~(\ref{eqn:array6})).
\vspace{1mm}
\noindent
\newline
$\bullet$ The earliest arrival time at node $j$ violates the time window constraint (refer Eq.~(\ref{eqn:array5}), Eq.~(\ref{eqn:array9}), Eq.~(\ref{eqn:array10})).
\vspace{1mm}
\noindent
\newline
$\bullet$ Node $j$ is a discharging station and the current SoC of the vehicle cannot support the complete trip from node $i$ to node $j$ and back to the depot; as well as the discharge operation at node $j$ refer Eq. ~(\ref{eqn:array7})).
\vspace{1mm}
\newline
\noindent
\textbf{Neural network architecture:} The architecture details and the hyperparameters used in the learning and testing are as follows: 
(i) architecture: $(5, 12, 6, 3, 1)$ consisting of 1 input and 1 output layer, and 3 hidden layers; 
(ii) optimizer: Adam; 
(iii) learning rate: $0.001$; 
(iv) batch size ($\beta$): $16$; 
(v) replay buffer size ($B$): $5000$; 
(vi) exploration policy: $\epsilon$-greedy with exploration factor decaying linearly from $1$ to $0$ over $75$ episodes.
The training consisted of $200$ episodes where each episode is a different instance (i.e., a random combination of customers, discharge stations, vehicles).
The neural network is realised using Pytorch library in Python~$3.6$.
The detailed training procedure is described in Algorithm~\ref{tab:rl-train}.
\vspace{1mm}
\newline
\noindent
\textbf{Neural network training:}
We train the RL agent using randomly generated datasets consisting of $20$\, customers; $5$\,discharging stations; $4$\,vehicles. 
The location coordinates of customers and discharging stations are generated uniformly within the range $\left[-100,100\right]$; while it is chosen uniformly random in the range $\left[-25,25\right]$ for the depot locations. 
The customer demand $c_{i}$ is drawn from an exponential distribution with $0.1$ scale parameter. 
The maximum loading capacity and battery SoC of each vehicle is chosen as $C = Q = 200$~units.
The speed of the vehicle, battery consumption rate $H$, and energy discharge rate $R$ at the discharging station is taken as $1$ unit each. 
The minimum time window $e_{i}$ is drawn randomly between $0$ and $200$~units, and the width of the time window ($l_{i}-e_{i}$) is chosen from a Gaussian distribution with mean of $35$~units and standard deviation of~$5$ units (minimum $1$ unit). For each training episode, a new random instance is generated using these parameters. The same parameters are then used for testing.
\vspace{1mm}
\newline
\noindent
\textbf{Solution improvement:} After a candidate solution is generated by the RL, it is further refined by a random insertion heuristic. 
The heuristic works by breaking up routes with the least number of customers, and each node in these routes is checked for feasibility by inserting them between various nodes in the remaining routes. 
Whenever such feasible locations are found, the one with the least marginal increase in the overall cost is chosen and the node is inserted to form a new route. 
This insertion heuristic, therefore, refines the solution by reducing the number of vehicles used; without having an adverse effect on the total cost of the trip.

\begin{algorithm}[h]
\caption{RL Training}
\label{tab:rl-train}
\DontPrintSemicolon
    Initialize the neural network with weights $\phi$\;
    Initialize batch size $\beta$, replay buffer $B$\;
    \For{episode = $1$ \KwTo TotalNumEpisodes}{
        Randomly choose data instance from training set\;
        Reset environment and get initial states\;
        \While{t < $T$}{
            Create a copy of the environment for local updates\;
            \While{free vehicle is unassigned}{
                Find feasible combinations of vehicle $u$, node $i$\; 
                \If{no feasible vehicle-node pair}{
                    break\;
                }
                Calculate $q_{t}(u,i)$= $\phi(S_{t}(u,i))$ $\forall$ ($u$, $i$) pairs\;
                Choose ($u$, $i$) pair using $\epsilon$-greedy assignment\;
                Perform local update on environment copy\;
            }
            Execute ($u$, $i$) assignments and get reward $R_{t}(u,i)$\;
            Add [$S_{t}(u,i)$; $R_{t}(u,i)$; $q_{t}(u,i)$] to replay buffer\;
        }
        Delete oldest entries in B if size exceeds buffer capacity\;
        Draw $\beta$ samples from B\;
        Update $\phi$ by minimizing MSE loss between $q_{t}(u, i)$ and $R_{t}(u,i)$ \;
    }
\end{algorithm}

\begin{table*}[t]
\caption{Performance comparison on Solomon datasets: GA vs. RL | \#C: num. customers | \#S: num. discharging stations | \\*\_d: distance travelled | *\_v: num. vehicles used | *\_ed: energy discharged | *\_t: compute time (seconds) | *\_cost: $\mathcal{M}$ (Eq.(1))}
\label{tab:ga-rl}
\begin{tabular}{@{}r|rrrrrrrrrrrr|r|r}
\toprule
\small \textbf{Dataset} & \small \textbf{\#C} & \small \textbf{\#S} & \small \textbf{GA\_d} & \small \textbf{RL\_d} & \small \textbf{GA\_v} & \small \textbf{RL\_v} & \small\textbf{GA\_ed} & \small \textbf{RL\_ed} & \small \textbf{GA\_t} & \small \textbf{RL\_t} & \small \textbf{GA\_cost} & \small \textbf{RL\_cost} & \small \textbf{GA > RL} & \small \textbf{\underline{GA_t}} \\
 &  &  &  &  &  &  &  &  &  &  &  &  & \textbf{cost (\%)} & \textbf{RL_t} \\
\midrule
CL1\_25 & 22 & 3 & 219.47 & 259.54 & 2.56 & 3.22 & 160.00 & 160.00 & 53.55 & 3.43 & 228.30 & 284.90 & 19.87 & 15.61\\
CL2\_25 & 22 & 3 & 209.58 & 281.73 & 1.75 & 3.00 & 67.50 & 168.75 & 68.62 & 3.50 & 168.86 & 230.5 & 26.74 & 19.6\\
RA1\_25 & 22 & 3 & 453.84 & 707.70 & 4.25 & 6.16 & 23.33 & 65.83 & 73.23 & 3.46 & 442.98 & 498.76 & 11.18 & 21.16\\
RA2\_25 & 22 & 3 & 368.86 & 696.39 & 2.09 & 3.72 & 15.46 & 104.55 & 79.74 & 3.54 & 222.10 & 299.10 & 25.74 & 22.52\\
RC1\_25 & 22 & 3 & 351.10 & 562.01 & 3.25 & 4.75 & 30.00 & 73.75 & 56.06 & 3.45 & 335.88 & 366.47 & 8.35 & 16.25\\
RC2\_25 & 22 & 3 & 322.09 & 595.21 & 2.13 & 3.50 & 23.75 & 136.25 & 82.61 & 3.50 & 221.86 & 250.69 & 11.50 & 23.6\\
\midrule
CL1\_50 & 45 & 5 & 431.08 & 607.83 & 5.00 & 6.22 & 360.00 & 510.00 & 180.38 & 8.66 & 435.10 & 496.46 & 12.36 & 20.83\\
CL2\_50 & 45 & 5 & 337.15 & 427.51 & 2.00 & 3.38 & 0.00 & 123.75 & 180.83 & 8.82 & 215.56 & 339.57 & 36.52 & 20.5\\
RA1\_50 & 45 & 5 & 797.99 & 1232.74 & 7.25 & 10.08 & 45.00 & 174.17 & 298.01 & 8.49 & 755.22 & 919.95 & 17.91 & 35.10\\
RA2\_50 & 45 & 5 & 628.19 & 1085.32 & 3.73 & 6.27 & 46.36 & 205.46 & 408.88 & 8.91 & 390.22 & 533.4 & 26.84 & 45.89\\
RC1\_50 & 45 & 5 & 718.44 & 1066.00 & 6.25 & 8.62 & 47.50 & 92.50 & 212.31 & 8.44 & 649.98 & 815.06 & 20.25 & 25.16\\
RC2\_50 & 45 & 5 & 602.01 & 1124.18 & 3.88 & 6.00 & 46.25 & 136.25 & 263.72 & 7.55 & 404.37 & 508.25 & 20.44 & 34.93\\
\midrule
CL1\_100 & 90 & 10 & 773.39 & 1089.53 & 9.00 & 11.22 & 160.00 & 270.00 & 307.42 & 31.45 & 904.02 & 1068 & 15.35 & 9.78\\
CL2\_100 & 90 & 10 & 545.19 & 751.99 & 3.00 & 4.50 & 0.00 & 33.75 & 269.99 & 34.37 & 324.73 & 461.23 & 29.60 & 7.86\\
RA1\_100 & 90 & 10 & 1253.89 & 1778.57 & 12.58 & 15.58 & 91.67 & 328.33 & 845.00 & 33.18 & 1302.78 & 1564.10 & 16.71 & 25.5\\
RA2\_100 & 90 & 10 & 925.22 & 1403.01 & 5.46 & 7.72 & 79.09 & 175.45 & 1466.74 & 33.74 & 568.48 & 715.07 & 20.50 & 42.22\\
RC1\_100 & 90 & 10 & 1438.64 & 2019.25 & 12.88 & 15.62 & 93.75 & 258.75 & 806.64 & 30.98 & 1338.50 & 1526.46 & 12.31 & 26.03\\
RC2\_100 & 90 & 10 & 1060.76 & 1636.08 & 6.63 & 9.13 & 92.50 & 233.75 & 1025.32 & 34.61 & 689.12 & 927.33 & 25.69 & 29.63\\
\bottomrule
\end{tabular}
\end{table*}
\vspace{-2mm}
\subsection{GA Formulation}

In this section, we will describe a GA metaheuristic constructed to solve the CEVRPTW-D problem. 
We will define a chromosome as a collection of routes of individual vehicles (also known as \emph{tour})~\citep{vaira2014GA}. 
Each chromosome $ch$ is assigned a fitness score using the following:
\begin{equation}\label{eq:fitness}
    f(ch) = Y_{1}.d_{1}(ch) + Y_{2}.v_{1}(ch) + w_{1}(ch).C_{1}(ch) - Y_{3}.E_d(ch)
\end{equation}
Here $f(ch)$, $d_{1}(ch)$, $v_{1}(ch)$, $C_{1}(ch)$ and $E_d(ch)$ denote the fitness score, total distance travelled, number of vehicles used, charge penalty, and total energy discharged back to the grid for a chromosome $ch$ respectively. 
The fitness score is chosen to align with Eq.(\ref{milp:obj}) with the exception of the term related to charge penalty $C_{1}(\cdot)$. 
This term ensures that the resulting solution does not violate the total charge constraints which ensures that vehicles don't get discharged before the end of the trip. 
Consequently, the weight $w_{1}(ch)$ is chosen to be a very high quantity ($w_{1}(ch) = 1000.d_{1}(ch)$). 
This guarantees that all chromosomes that violate the total charge constraints are not chosen in the next generation. 
In addition, for every chromosome, we delete all routes that comprise of only discharging station visits with zero customer visits. The objective is to minimize the fitness score; and hence, chromosomes with lower fitness score are considered ``better''.
The GA proceeds according to the following steps.
\vspace{1mm}
\newline
\noindent
$\bullet$ \emph{Initial Population Generation:} It is done via the nearest neighbor rule starting from random initial nodes, and respecting the total distance and time window constraints.
\vspace{1mm}
\newline
\noindent
$\bullet$ \emph{Initial Solution Improvement:} An insertion heuristic is used in this step.
It involves breaking up routes with least volume utilization; putting their nodes in an unreserved pool; trying to insert these nodes into existing routes respecting constraints; and finally choosing the newly formed route with the smallest increment in fitness score. 
This reduces the number of vehicles in the solution.
\vspace{1mm}
\newline
\noindent
$\bullet$ \emph{Selection of Parents:} A binary tournament procedure is followed for parent selection wherein two chromosomes are selected at random; and the one with lower fitness score is chosen.
\vspace{1mm}
\newline
\noindent
$\bullet$ \emph{Crossover and Mutation:} Two parents are crossed over either using a \emph{common nodes crossover} or a \emph{common arcs crossover} approach to produce new offsprings. 
Once the new offsprings are created, they undergo two stages of mutation. 
The first stage involves creation of new members by removing routes with only discharging stations (and no customer visits). 
The second stage involves a mutation with $10\%$ probability (in this case, the type of mutation is chosen to be one among several with equal likelihood; for e.g., random note/route removal, nearest node removal).
\vspace{1mm}
\newline
\noindent
$\bullet$ \emph{Progression and Termination:} The size of the population at the beginning of each generation is a fixed quantity (e.g., $200$). 
Once the parents and offsprings are created and added to the population of current generation; we follow an elitist procedure to select the population for the next generation. 
In this method, we retain the top $10\%$ of the current population; while the others are randomly sampled. 
This procedure is terminated if there is no improvement in the solution, or a generation count limit is reached.

\section{Evaluation} \label{sec:evaluation}

\begin{table*}[t]
\caption{Performance comparison on specific instances of Solomon datasets: MILP vs. GA vs. RL}
\label{tab:milp-ga-rl}
\footnotesize
\begin{tabular}{@{}r|rrrccccccrrrrrr@{}}
\toprule
\textbf{DataSet} & \textbf{MILP\_d} & \textbf{GA\_d}  & \textbf{RL\_d}  & \textbf{MILP\_v} & \textbf{GA\_v} & \textbf{RL\_v} & \textbf{MILP\_ed} & \textbf{GA\_ed} & \textbf{RL\_ed} & \textbf{MILP\_t} & \textbf{GA\_t}  & \textbf{RL\_t} & \textbf{MILP\_cost} & \textbf{GA\_cost}   & \textbf{RL\_cost}    \\ 
\midrule
CL101            & 214.56           & 214.71         & 250.22         & 3                                    & 3                                  & 3                                  & 180                                   & 180                                 & 90                                  & 7068             & 68.10          & 3.46           & 268.42              & 268.43            & 291.99            \\
CL201            & 218.60           & 219.58         & 290.56         & 2                                    & 2                                  & 3                                  & 90                                    & 90                                  & 270                                 & 3029             & 64.81          & 3.52           & 189.06              & 189.10            & 248.81            \\
RA105            & 620.44           & 556.81         & 632.62         & 5                                    & 5                                  & 5                                  & 40                                    & 30                                  & 20                                  & 1078             & 69.58          & 3.53           & 521.10              & 521.33            & 526.49            \\
RA109            & 504.19           & 460.52         & 634.91         & 4                                    & 4                                  & 5                                  & 40                                    & 30                                  & 30                                  & 31417            & 68.41          & 3.48           & 415.18              & 416.11            & 524.09            \\
RC101            & 478.56           & 462.15         & 488.604        & 4                                    & 4                                  & 4                                  & 40                                    & 30                                  & 20                                  & 2608             & 65.39          & 3.56           & 414.27              & 416.17            & 419.58            \\
RC106            & 346.23           & 346.50         & 367.23         & 3                                    & 3                                  & 3                                  & 30                                    & 30                                  & 20                                  & 7308             & 45.01          & 3.42           & 310.25              & 310.26            & 313.47            \\
RC102            & 352.65           & 352.74         & 368.17         & 3                                    & 3                                  & 3                                  & 30                                    & 30                                  & 20                                  & 89857            & 43.46          & 3.44           & 310.48              & 310.48            & 313.50            \\
RC105            & 465.09           & 412.37         & 489.84         & 4                                    & 4                                  & 4                                  & 40                                    & 30                                  & 20                                  & 89835            & 88.47          & 3.53           & 413.80              & 414.40            & 419.62            \\ 
\bottomrule
\end{tabular}
\end{table*}
To test the performance of the aforementioned algorithms, we use the Solomon datasets as benchmarks~\citep{solomon1987}, which are modified to suit the EVRPTW-D problem. These datasets are classified into two categories: Type~$1$, (vehicles with low capacities); and Type~$2$ (vehicles with larger capacities). Further, within each type the dataset can have customer locations that are either clustered $(CL)$, random $(RA)$, or a combination of both $(RC)$. 
For each standard dataset, we convert some of the existing customers to charging stations with a pre-specified timing window corresponding to the grid peak demand. 
For instance, the standard dataset $CL1_{100}$ comprises of $90$~customers and $10$~discharging stations.
\newline
\indent
The experiment results are captured in Tables~\ref{tab:ga-rl}~and~\ref{tab:milp-ga-rl}. 
Each row, respectively, describes the average results obtained for datasets of a specific type ($1$ or $2$); number of customers and discharging stations ($25$, $50$, $100$); and location distribution of customers ($CL$, $RA$, $RC$).
We make comparisons using the objective value $\mathcal{M}$ (refer to Eq.~\ref{milp:obj}), which is averaged over all datasets corresponding to a particular row.
Table~\ref{tab:ga-rl} outlines the comparison results between the RL approach of \quik\ and GA for all the datasets. 
Table~\ref{tab:milp-ga-rl} contains specific instances where the MILP formulation (Section~\ref{sec:milp}) converges within a reasonable amount of time.
\newline
\indent
From Table~\ref{tab:milp-ga-rl}, we can observe that GA metaheuristic performs close to an exact formulation approach\footnote{The MILP approach was solved using the standard Coin-BC (Branch and Cut ) Solver}. Table~\ref{tab:ga-rl} shows two key aspects: first, GA outperforms RL by an average of $19.8\%$ (range $8.3\%-36.52\%$) based on $\mathcal{M}$; and second, RL is faster than GA by $24$ times (on average). 
In other words, the RL approach of \quik\ finds approximately optimal solutions at a much faster rate than the metaheuristic GA. The analysis also reveals that the average optimality gap between GA and RL is consistent with respect to number of nodes:  $17.23\%$ (std. dev. of $7.98\%$) for $25$-nodes; $22.38\%$ (std. dev. of $8.35\%$) for $50$-nodes; and $20.02\%$ (std. dev. of $6.58\%$) for $100$-nodes. Therefore, for datasets with higher number of nodes (prevalent in real-world scenarios), \quik\ can be preferred over GA (if compute time is critical).

\section{Related Work} \label{sec:relatedwork}

Electric VRP or EVRP is one of the latest variations of the extensively studied vehicle routing problem.
It was first introduced by Schneider et al.~\citep{schneider2014:evrptw} where they explored the possibility of recharging EVs during the trip while considering the capacity constraints of the vehicle and time windows for goods delivery.
It was an extension of the work on green VRP by Erdogan and Miller~\citep{erdogan2012:green-vrp}, which aimed to find a way to extend the range of alternate fuel vehicles (i.e., those running on biodiesel, liquid natural gas, or CNG) by visiting refueling stations.
However, the key difference between these two works is the energy replenishment model; wherein~\citep{erdogan2012:green-vrp} uses a constant replenishment time model, while~\citep{schneider2014:evrptw} uses a linear charging time model. 
Following these two initial studies, there have been multiple investigations on EV route optimization that extend EVRP to include various operational aspects of charging and discharging. 
Existing work can be broadly categorized as follows:
\vspace{1mm}
\newline
\noindent
(i) \emph{Grid-to-Vehicle} (G$2$V):
Since EV charging is time consuming, Felipe et al.~\citep{felipe2014:greenVRP} and Keskin et al.~\citep{keskin2016:partialcharging} proposed partial recharging strategies rather than going for full charge.
The key challenge in this EVRPTW-PR problem was to estimate the charge levels for a large set of EVs using different types of charging technologies; for which ~\citep{felipe2014:greenVRP} proposed simulated annealing metaheuristic approach combined with a constructive local search mechanism, while~\citep{keskin2016:partialcharging} developed an adaptive large neighborhood (ALNS) search method.
Yang et al.~\citep{yang2015:evrp-tou} factored in the time-of-use electricity price for estimating the charging cost with the aim to optimize the EV route cost for delivery and pickup services. 
They solved this model by developing a learnable partheno-genetic algorithm with
integration of expert knowledge about location of charging stations and their selection probability.
Barco et al.~\citep{barco2017:ev} extended this problem of route assignment and charging actions to reduce the combined cost of charging and battery degradation, and solved using the differential evolution technique.
Yu et al.~\citep{yu2018:avls} developed an MILP model to find optimal routes for autonomous EVs in order to make them charge at locations of energy generation.
\vspace{1mm}
\newline
\noindent
(ii) \emph{Vehicle-to-Grid} (V$2$G) \& \emph{Grid-to-Vehicle} (G$2$V):
Tang et al.~\citep{tang2017:ev-smartgrid} proposed a distributed routing and charge scheduling algorithm, which decides the charging station locations and the charge / discharge rates along the optimal route for EVs to complete their trip by incurring the least travel cost.
In this problem, they considered two types of stations with time invariant energy price signals: one, providing renewable energy at a low price; and second, regular stations with higher charging cost with rewards for discharging.
Trivino et al.~\citep{trivino2019:ev-smargrid} extended this work by incorporating intermediate stops for EVs, and considering time-variant electricity prices and battery degradation cost.
Abdulaal et al.~\citep{abdulaal2017:multivariant-EVRP} studied G$2$V and V$2$G options in the travel path of EVs with multiple constraints, and developed a GA solver that incorporates Markov decision process and trust region optimization methods.
Lin et al.~\citep{lin2021:evrp} studied the problem of joint optimization of routing and charge/discharge scheduling of multiple EVs under time-variant electricity prices; for which they proposed a MILP solution with hueristic.
\newline
\indent
The application of RL to EVRP variants is relatively new. Lin et al.~\citep{lin2020:rl-evrp} proposed a RL framework to solve capacitated EVRP with time window constraints (CEVRPTW).
This paper differs from~\citep{lin2020:rl-evrp}, both in terms of the problem definition and RL design.
Our multi-service model of delivering goods and energy requires extending CEVRPTW with energy discharge constraints on account of visiting discharging stations during a delivery trip.
We use a RL model to find (near) optimal routes by mapping vehicles to customer / discharging station locations; instead of using a RL attention mechanism to handle dynamic changes in the operating constraints. 

\section{Conclusion} \label{sec:conclusion}

In this paper, we presented an approach for the solving CVRPTW-D (a new EVRP variant) using a RL framework.
Using numerical simulations on benchmark datasets, we showed that the RL approach generates solutions $24$ times faster than the proposed GA and MILP baselines; although with approximately $20$\% increase in optimality gap in terms of the cost.
We are currently working on designing better RL models to not only close this optimality gap, but also to handle dynamic demands in real-time.


\balance
\bibliographystyle{ACM-Reference-Format}
\bibliography{evrp_references} 


\end{document}